\documentclass[a4paper]{article}

\usepackage{INTERSPEECH2021}
\usepackage{multirow}
\usepackage{multicol}
\usepackage{cases}
\usepackage{array}
\usepackage{amsmath}
\usepackage{color}
\usepackage{xcolor}

\title{Multilingual Speech Evaluation: \\Case Studies on English, Malay and Tamil}

\name{Huayun Zhang, Ke Shi, Nancy F. Chen}
\address{Institute for Infocomm Research, A*STAR, Singapore}
\email{zhang\_huayun, shi\_ke, nfychen@i2r.a-star.edu.sg}

\begin{document}

\maketitle
\begin{abstract}
 Speech evaluation is an essential component in computer-assisted language learning (CALL). While speech evaluation on English has been popular, automatic speech scoring on low resource languages remains challenging. Work in this area has focused on monolingual specific designs and handcrafted features stemming from resource-rich languages like English. Such approaches are often difficult to generalize to other languages, especially if we also want to consider suprasegmental qualities such as rhythm. In this work, we examine three different languages that possess distinct rhythm patterns: English (stress-timed), Malay (syllable-timed), and Tamil (mora-timed). We exploit robust feature representations inspired by music processing and vector representation learning. Empirical validations show consistent gains for all three languages when predicting pronunciation, rhythm and intonation performance.

\end{abstract}


\section{Introduction}
Learning online has become the new norm after the COVID-19 pandemic. Increasing demand for online language learning has made Computer-Aided Pronunciation Training (CAPT) even more popular than before. Automatic speech evaluation, a key component in CAPT, aims to score speaking proficiency according to the standardized assessment criteria \cite{levy2013call,ESKENAZI2009832, 7820782}.

Traditional speech assessment studies usually focus on extracting features from the Hidden Markov Model (HMM) based automatic speech recognizer (ASR) (e.g. \cite{596227,WITT200095,4518800}).
\cite{596227} explored posterior probabilities and duration related features in automatic scoring. As a variation of likelihood ratio, Goodness of Pronunciation (GOP) \cite{WITT200095} is one of the the most widely adopted feature in speech evaluation task \cite{4518800}. With the development of deep neural network (DNN), GOP was further optimized to predict better phone segmentation and posterior estimation \cite{hu2013new,HU2015154,huang2017transfer}. Weighted GOP was also proposed to improve its discriminative ability in non-native speech \cite{van2010using}. Recently, an ASR-free approach was proposed, which drives features from the marginal distribution of speech signals \cite{cheng2020asrfree}.

Linguistic features have also attracted research interests. As in \cite{8639697}, a prompt-aware feature was proposed for spontaneous speech evaluation. A context-aware GOP was proposed in\cite{shi2020context} to incorporate phone transition factor and phone duration factor into the calculation of GOP. \cite{7404814} leveraged Bidirectional Long Short Term Memory (Bi-LSTM) to learn high-level abstraction features by encoding both time-sequence and time aggregated information from speech. \cite{8462562} proposed to encode lexical information and acoustic information in separate neural networks. All these studies are focused on languages with rich resources. Extracting cross language effective features, especially on low resource languages, is still challenging.

In terms of modeling approaches, early scoring strategies were based on statistical models such as Gaussian process \cite{van2015automatically}. Recent studies employed more deep learning approaches. \cite{lee2016language} proposed to utilize Convolution Neural Network (CNN) along with a multi-layer perceptron classifier. \cite{tao2016exploring} explored three deep learning based acoustic models including Tandem GMM-HMM, DNN,  CNN, and found they provide substantial improvement in scoring performance. Long short-term memory recurrent network (LSTM) was adopted in pronunciation assessment \cite{yu2015using,Li2017}. More recently, attention mechanism has also been applied \cite{lin2020automatic,8462562,8639697} to speech evaluation. 
These studies have presented promising improvement on speech evaluation performance in the language specific tasks. However, multilingual speech scoring task was not explored.

In this paper, we propose a unified framework for fine-grained multilingual speech evaluation on assessing pronunciation, rhythm, and intonation by leveraging robust feature representations. Specifically, we investigate rhythm-aware tempo features and multilingual vector representations. Moreover, multi-task learning strategy is employed to further improve the evaluation performance on the low-resource languages.

\section{Methods}
\subsection{Setup}
\begin{figure}[ht]
    \centering
    \includegraphics[width=0.35\textwidth]{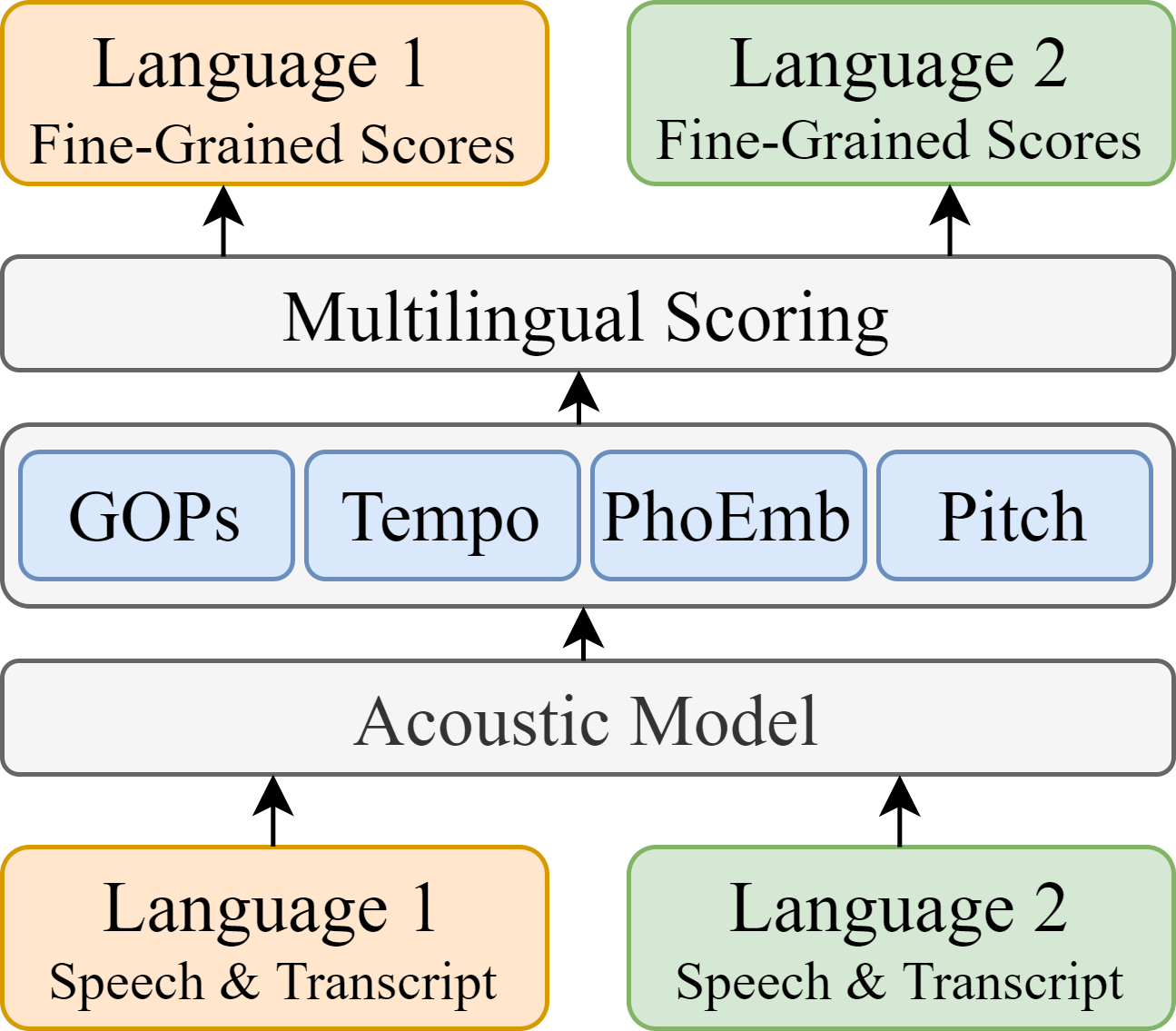}
    \caption{Speech evaluation framework, fine-grained scores include the metrics of pronunciation, fluency, and intonation.}
    \label{fig:system}
\end{figure}

All the experiments in this paper follow the system diagram in Figure \ref{fig:system}: Given a speech utterance from a student, forced alignment and phone decoding is applied to an acoustic model to obtain phoneme level timing and the acoustic log likelihood. Various features are derived to train scoring model to predict fine-grained scores to quantify oral language skills (i.e. pronunciation, rhythm, and intonation).\\

\noindent\textbf{Performance Metrics}: The Pearson Correlation Coefficient (PCC) and Mean Square Error (MSE) between model prediction and the average score of teachers.

\subsection{Acoustic Model}
Mel-filter bank feature (29 dimensional) was adopted in acoustic modeling. Optimized on various ASR tasks, this model is designed as a combination of different neural networks: At the bottom levels, a stack of specially designed convolution neural network (CNN) running on 2D windows across time and frequency is trained to extract robust intermediate representation from the filter bank outputs. 
Time Delay Neural Network (TDNN) and LSTM are placed on top of these CNN. The output of the final LSTM is linked to 1500 output senones via a fully connected layer. This model was trained on WSJ \cite{paul-baker-1992-design}, Switchboard \cite{225858} and Fisher \cite{WITT200095}, using lattice-free Maximum Mutual Information (MMI) criterion with a sub-sampling factor of 3 \cite{Hadian2018}. 
To minimize the acoustic mismatch on L2 speech evaluation tasks, acoustic adaptation was implemented using the speechocean762 training data as described in Section 3.\\

\noindent\textbf{Malay and Tamil Acoustic Models:} Malay and Tamil acoustic models have the same configuration and were trained on 600 hours' Malay speech (1,500 native speakers from Singapore and Malaysia) and 200 hours' Tamil speech (700 native speakers from Singapore) respectively.

\subsection{Feature Representation}
\label{sec:feature}
\subsubsection{Acoustic Posterior Probability}
Goodness of Pronunciation (GOP) \cite{WITT200095} has been widely adopted in speech evaluation. Phone level GOP is the time average of log posterior probability over the phone duration:
\begin{equation}
\begin{aligned}
\small
    \mathrm{GOP}(p) =\frac{1}{T}log(P(p|\mathbf{O}))\\
    \end{aligned}
\end{equation}
where $\mathbf{O}=[\mathbf{o}_1,...,\mathbf{o}_T]^T$ is a speech segment of phone $p$ in the alignment.

\begin{equation}
\begin{aligned}
\small
     &P(p|\mathbf{O})=\frac{P(p)P(\mathbf{O}|p)}{\sum_{q\in{Q}}{P(q)P(\mathbf{O}|q)}}\\
    \end{aligned}
\end{equation}
where $Q$ is the collection of all possible paths on the observation $\mathbf{O}$. $P(p)$ and $P(q)$ are the priors of $p$ and $q$.

\subsubsection{Tempo / Duration}
In musical terminology, tempo is the pace of a given piece \cite{scheirer1998tempo}. It is usually measured in beats per minute. We borrow this concept to measure the phonological timing patterns in speech. In this study, speech tempo is defined as a combination of speaking rate and normalized phone duration:
\begin{equation}
   \mathrm{T}(p) = \mathrm{concat}(1/\tau, (\tau - \mu)/\sigma
   )
\end{equation}
where $\tau$ is the duration of current canonical phone in the alignment, $(\mu,\sigma)$ are the normal distribution parameters of $\tau$ in the sentence.
The instant phone tempo is spliced with its context into a tempo vector:
\begin{equation}
   \mathbf{T}(p) = \mathrm{concat}(\mathrm{T}(p_{i-k}),....,\mathrm{T}(p_{i}),...,\mathrm{T}(p_{i+k}))
\end{equation}
where $k$ is the number of neighboring phones considered in each direction.\\

For comparison, phone duration feature was also tested. Similar to the tempo feature, phone duration and duration difference between successive phones are spliced with its neighbors into a duration vector.

\subsubsection{Multilingual Phonemic Embedding}



In this work, we propose to use a multilingual vector representation to characterize the spoken utterances from English, Malay and Tamil. Representing the speech signal with phonetic features from other languages have shown to be useful in many tasks, including speech recognition \cite{zissman1996comparison}, spoken term detection \cite{wallace2007phonetic}, speech summarization \cite{chen2013minimal}, and spoken language  identification \cite{pellegrino2000automatic}. However, such multilingual representations have not been applied to speech evaluation.
Therefore, in this work, we adapt distributed linguistic representation and apply well tested approaches in NLP to characterize the multilingual phonemic space. A phoneme embedding matrix was estimated by Google's Word2Vec \cite{41224} on a multilingual training corpus (words were mapped to phoneme strings). Items in language-specific phoneme tables $Q_{English}$,$Q_{Malay}$ and $Q_{Tamil}$ are prefixed with corresponding language ID and merged together to form a multilingual phoneme table,$Q_{Multilingual}$. Each canonical phone in alignment is assigned a unique one-hot index from this multilingual phoneme table. By multiplying the embedding matrix, this one-hot phoneme index is transformed into a $D$ dimensional vector. $D=32$ in this study.

\subsubsection{Pitch}
Pitch provides acoustic cues for a speaker's intonation, confidence and expressiveness. In this study, a feature vector including raw log pitch, normalized log pitch, delta log pitch and wrapped NCCF (Normalized Cross Correlation Function) was extracted from Kaldi \cite{6854049}. Frame-wise pitch vectors are averaged in each canonical phone's duration.

\subsubsection{Feature Assembly}



For each canonical phone, relevant features are concatenated in a sequence.
\begin{equation}
\footnotesize
\mathbf{V}(p)=\{\mathrm{GOP}(p),\mathrm{Tempo}(p),\mathrm{PhoEmb},\mathrm{Pitch}\}
\end{equation}
At the end of each sentence, an utterance ending symbol is appended.

\subsection{Scoring Module}
The scoring module aims to map the meta features described in Section \ref{sec:feature} to the fine-grained proficiency scores assigned by human raters. RNN is investigated as the deep learning backbone, in particular, stacked Bi-directional long short term memory (Bi-LSTM) is utilized in sequential modeling. Finally, the hidden representations encoded by RNN are fed into a linear layer followed by a $\mathrm{tanh}$ activation function to predict the scores.

The human scores are re-scaled into the range of $(-1,+1)$ before training, and the model predictions are scaled back before comparing with human scores. MSE is chosen as the loss function in training. In this study, both monolingual and bilingual models have approximately the same numbers of parameters (around 2M).

\subsection{Low Resource Speech Evaluation}
\label{low-resource}
Collecting and labeling L2 speech data is time consuming and labor intensive, so data preparation is often the bottleneck when developing speech evaluation models for low resource languages. From our experience, we speculate that for the same type of fine-grained metrics, be it pronunciation, rhythm, or intonation, there is certain language agnostic information that teachers use during scoring. This type of language agnostic information could help mitigate the adverse effects stemming from the scarcity of linguistic resources.

In Singapore, while Malay and Tamil are two of the four official languages and there is strong support to develop spoken language technology to help students learn their ethnic mother tongues, 67\% Malay households and 70\% Indian households speak English as their main language. Only 3\% of the population speaks Tamil at home\cite{household}.
Therefore, we employ two strategies to tackle the low-resource challenge. The first is model adaptation: Tamil model could be adapted from a well trained English or Malay model. The second is data augmentation: Multilingual tasks are learned simultaneously by sharing most model parameters and a language-specific linear layer before the output $\mathrm{tanh}$.

In this work, we used Malay and Tamil for multilingual speech evaluation as the teaching scoring data are more homogeneous; both Malay and Tamil were scored by teachers certified by the Ministry of Education in Singapore, whereas the English data was scored outside Singapore. 

\begin{table*}[htp!]
\linespread{1}
\centering
\small
\begin{tabular}{l|cc|cc|cc}
\hline
  \multirow{2}*{\textbf{Feature}} &  \multicolumn{2}{c|}{\textbf{Pronunciation}} & \multicolumn{2}{c|}{\textbf{Rhythm}} & \multicolumn{2}{c}{\textbf{Intonation}} \\
\cline{2-7}
 ~ & \textbf{MSE} & \textbf{PCC} & \textbf{MSE} & \textbf{PCC} & \textbf{MSE} & \textbf{PCC}\\
\hline
\textit{GOP} &1.428 & 0.639 & 1.065 & 0.694 & 1.009 & 0.713 \\
\textit{GOP+Dur} & 1.392 & 0.647 & 1.021 & 0.707 & 1.018 & 0.710 \\
\textit{GOP+Tempo}  & 1.346 & 0.662 & 1.006 & 0.712 & 0.980 & 0.723  \\
\textit{GOP+Tempo+PhoEmb}  & \textbf{1.341} & \textbf{0.667}  & \textbf{0.958} & \textbf{0.729} & \textbf{0.970} & \textbf{0.726} \\
\textit{GOP+Tempo+PhoEmb+Pitch}  & 1.368 & 0.654 & 1.037 & 0.702 & 1.064 & 0.699 \\
\hline
\textit{Human} & - & 0.754 & - & 0.767 & - & 0.753 \\
\hline
\end{tabular}
\caption{Results on English: MSE and PCC scores of different model and feature configurations. }
\label{tab:English_results}
\vspace{-0.2cm}
\end{table*}


\begin{table*}[ht]
\linespread{1}
\centering
\small
\begin{tabular}{p{3.39cm}|p{3.98cm}|cc|cc|cc}
\hline
\multirow{2}*{\textbf{Model}} &
 \multirow{2}*{\textbf{Feature}} &\multicolumn{2}{c|}{\textbf{Pronunciation}} & \multicolumn{2}{c|}{\textbf{Fluency}} & \multicolumn{2}{c}{\textbf{Intonation}} \\
\cline{3-8}
 ~&~ & \textbf{MSE} & \textbf{PCC} & \textbf{MSE} & \textbf{PCC} & \textbf{MSE} & \textbf{PCC}\\
\hline
\multirow{5}*{Monolingual}&\textit{GOP}& 0.537 & 0.473 & 0.580 & 0.487 & 0.707 & 0.415\\
&\textit{GOP+Dur}& 0.536 & 0.474 & 0.579 & 0.487 & 0.703 & 0.421\\
&\textit{GOP+Tempo}& 0.530 & 0.482 & 0.572 & 0.497 & 0.697 & 0.429 \\
&\textit{GOP+Tempo+PhoEmb} & 0.523 & 0.494 & 0.563  & 0.509 & 0.691  & 0.437  \\
&\textit{GOP+Tempo+PhoEmb+Pitch} & 0.518 & 0.500 & 0.546  & 0.531 & 0.658  & 0.479 \\
\hline
\textit{Malay, Tamil Bilingual }&\textit{GOP+Tempo+PhoEmb+Pitch} & \textbf{0.506} & \textbf{0.518} & \textbf{0.538}  & \textbf{0.540} & \textbf{0.656}  & \textbf{0.482} \\
\hline
\textit{Human} & & - & 0.547 &- & 0.571 &- & 0.545  \\
\hline
\end{tabular}
\caption{Results on Malay: MSE and PCC scores of different feature configurations.}
\label{tab:Malay_results}
\vspace{-0.2cm}
\end{table*}

\begin{table*}[htp!]
\linespread{1}
\centering
\small
\begin{tabular}{l|l|cc|cc|cc}
\hline
   \multirow{2}*{\textbf{Model}} & \multirow{2}*{\textbf{Feature}} & \multicolumn{2}{c|}{\textbf{Pronunciation}} & \multicolumn{2}{c|}{\textbf{Fluency}} & \multicolumn{2}{c}{\textbf{Intonation}} \\
\cline{3-8}
 ~ &~ & \textbf{MSE} & \textbf{PCC} & \textbf{MSE} & \textbf{PCC} & \textbf{MSE} & \textbf{PCC}\\
\hline
\multirow{5}*{Monolingual}&\textit{GOP}& 0.344 & 0.490 & 0.354 & 0.584 & 0.259 & 0.552 \\
&\textit{GOP+Dur}& 0.334 & 0.513 & 0.348 & 0.594 & 0.254 & 0.563 \\
&\textit{GOP+Tempo}& 0.333 & 0.516 & 0.342 & 0.603 & 0.250 & 0.573 \\
&\textit{GOP+Tempo+PhoEmb}& 0.330 & 0.522 & 0.348 & 0.594 & 0.260 & 0.552\\
&\textit{GOP+Tempo+PhoEmb+Pitch} & 0.332 & 0.518 & 0.347  & 0.595 & 0.258  & 0.555 \\
\cline{1-8}
\textit{Monolingual (Adaptation)}&\textit{GOP+Tempo+PhoEmb+Pitch } & 0.334 & 0.519 & 0.339  & 0.608 & 0.254  & 0.562 \\
\cline{1-8}
\textit{Tamil, Malay Bilingual}&\textit{GOP+Tempo+PhoEmb+Pitch} & \textbf{0.324} & \textbf{0.534} & \textbf{0.338}  & \textbf{0.608} & \textbf{0.249}  & \textbf{0.575} \\
\hline
 \textit{Human}&  & - & 0.568 & - & 0.619 & - & 0.582  \\
\hline
\end{tabular}
\caption{Results on Tamil: MSE and PCC scores of different feature configurations and training strategies.}
\label{tab:Tamil_results}
\vspace{-0.5cm}
\end{table*}

\section{Experiments}
\subsection{Speech Corpora}
\label{sec:corpora}
Experiments were conducted in three languages: English, Malay, and Tamil. \\

\noindent\textbf{English:}
 Speechocean762\footnote{http://www.openslr.org/101/}, a recently released data set for speech evaluation was investigated in this study. It consists of 5000 English utterances collected from 250 nonnative speakers. Half of the data, i.e. 2500 utterances and 125 speakers are reserved as the test data. There is no speaker overlap between testing and training. Mandarin Chinese is the first language for all speakers. Half of the speakers are children. For each utterance, five raters' scores are provided at 3 levels: phoneme level, word level and sentence level. Sentence level scores were investigated in this study. The average inter-rater PCC are 0.754, 0.767and 0.753 on pronunciation, rhythm, and intonation respectively.\\

\noindent\textbf{Malay:}
 Our Malay corpus contains 14,088 utterances and 230 Singapore speakers between 9-16 years old. The average inter-rater PCC are 0.547, 0.571, 0.545 on pronunciation, fluency, and intonation respectively.\\

\noindent\textbf{Tamil:}
 Our Tamil corpus consists of 5,215 utterances collected from 100 Singapore speakers between 9-16 years old. Each utterance was scored by four raters. The average inter-rater PCC are 0.568, 0.619, and 0.582 on pronunciation, fluency, and intonation respectively.\\

\noindent\textbf{Score Annotations from Teachers:}
English utterances from Speechocean762 are scored by human-raters independently using a 10-point scale (1 is the lowest, 10 is the highest). Malay and Tamil utterances were scored by human-raters independently using a 5-point scale. (1 is the lowest, 5 is the highest). For Malay and Tamil, the average rating scores were used as ground truth scores.
For each corpus, multiple inter-rater PCC were calculated between the scores of one rater and the average scores of the rest of all raters \cite{lin2020automatic}. By averaging all inter-rater PCC, the upper bound of the scoring performance (Human performance) was obtained (see the bottom lines in Table \ref{tab:Malay_results}-\ref{tab:Tamil_results}). For Speechocean762, the median scores were adopted following the example score files coming with the database. Multiple inter-rater PCC were calculated between the scores of one rater and the median scores of the rest of all raters as shown in Table \ref{tab:English_results}.

\subsection{Results}
All model and feature configurations are compared with PCC and MSE metrics following setup in previous work \cite{8462562,lin2020automatic}. \\
\\
\noindent\textbf{English Experimental Results}:
The results for English are shown in Table \ref{tab:English_results}\footnote{Speechocean762's scoring categories are accuracy, fluency and prosody, which primarily evaluate the pronunciation, rhythm and intonation according to http://www.openslr.org/101/. We adapt the terminology to be more consistent with other datasets to make it easier for comparing performance.}. It is expected that GOP feature performs acceptably on the speech evaluation task. The scoring performance was improved by replacing normal duration feature with tempo feature, which reduced MSE by 3.3\%, 1.5\%, 3.7\% relatively and improved PCC by 2.3\%,0.7\%, 1.8\% relatively on the three oral proficiency measures, i.e. pronunciation, rhythm, and intonation, respectively.
The phoneme embedding feature boosted the performance further by 0.4\%, 4.8\%, 1.0\% relative MSE decrements and 0.8\%, 2.4\%, 0.4\% relative PCC improvements on the three oral proficiency measures.\\
\\
\noindent\textbf{Malay Experimental Results:}
The results for Malay are shown in Table \ref{tab:Malay_results}. Replacing duration feature with tempo feature brought 1.1\%, 1.2\% and 0.9\% relative MSE reductions and 1.7\%, 2.1\% and 1.9\% relative PCC improvements on the three oral proficiency measures respectively. By using phoneme embedding feature, scoring performance was further improved by 1.3\%, 1.5\% and 0.9\% relatively for MSE and 2.5\%, 2.4\% and 1.9\% relatively for PCC on the three oral proficiency measures respectively. Pitch feature improved the PCC performance by another 1.2\%,4.3\% and 9.6\% relatively. In addition, multi-task learning strategy benefited to Malay speech scoring task, especially for pronunciation and fluency proficiency measures. \\
 \\
\noindent\textbf{Tamil Experimental Results:}
Table \ref{tab:Tamil_results} shows the results for Tamil. Similar to English and Malay, the tempo feature performs better than duration. The multilingual embedding feature brought improvements on pronunciation while the pitch feature brought improvements on fluency and intonation. As data scarcity is a major bottleneck for Tamil speech evaluation, two cross language training strategies were further investigated: acoustic adaptation and data augmentation (multi-task learning). The results in Table \ref{tab:Tamil_results} show that both methods are effective. Especially, multi-task learning reduced MSE by 2.4\%, 2.6\%, 3.5\% relatively and improved PCC performance by 3.1\%, 2.2\%, and 3.6\% relatively on the three oral proficiency measures respectively.

\section{Discussion}

\noindent\textbf{Speech Tempo:}
There have been studies on speech measurements that compare different rhythmic patterns across languages \cite{RAMUS2000AD3,inbook,ling2000q}. We adopted speech tempo as the three languages we are investigating are known to possess distinct rhythm patterns: stress-timed for English, syllable-timed for Malay and mora-timed for Tamil. We empirically show that speech tempo features are straightforward to use and effective in speech evaluation modeling, showing consistent improvements compared to traditional duration features for the PCC metric.


\noindent\textbf{Multilingual Phoneme-Aware Scoring:}
By introducing a multilingual phoneme embedding feature, data from the same phoneme is trained in a phoneme-specific subspace, while data from different phonemes would be trained in separate sub-spaces.
The variability in the scoring model is attributed to both the pronunciation variation that is phoneme independent and the pronunciation variation that is phoneme dependent. Phoneme aware modeling decouples these two variations and provides a better prediction on the language proficiency.

\noindent\textbf{Cross Lingual Modeling:}
On low resource tasks such as Tamil, we attempted to improve model performance by leveraging from Malay. Both model adaptation and data augmentation have been shown to be effective. Consistent improvement was observed, suggesting language-agnostic information could potentially help speech evaluation scoring. 
Especially data augmentation improved data efficiency and alleviated training data over-fitting on low resource speech scoring tasks. 
We did explore using English data and models to help Tamil, but only observed minimal gains. We suspect this is because the English data was scored from a different standard than Malay and Tamil (the latter two is based on needs of Singapore Education). How to further exploit English resources to help lower-resource languages for speech evaluation is a theme of on-going work.

\noindent\textbf{Pitch:}
 In the Malay task, We observed obvious contribution by using pitch feature though similar trends were not observed for other languages such as English. We leave further analysis and investigations on how to more appropriately exploit pitch-related features for future work.

\vspace{-0.2cm}
\section{Conclusion}
We systematically compared different feature configurations on multilingual speech evaluation tasks, focusing on sentence level fine-grained metrics. Tempo feature and multilingual phoneme embedding features were introduced. Consistent improvements were observed in experiments by adopting tempo-aware and phoneme-aware features in evaluation modeling. While Malay and Tamil are from different language, cross-lingual experiments showed that data and models in other languages could help improve speech evaluation performance.
In the future, we will explore unsupervised error pattern discovery to diagnose speaker-specific pronunciation problems \cite{7472858}.



\bibliographystyle{IEEEtran}

\bibliography{IS2021}

\end{document}